\documentclass[sigconf]{acmart}

\usepackage{bm}
\usepackage{threeparttable}
\usepackage{multirow}
\usepackage{makecell}
\usepackage{balance}

\usepackage{setspace}

\AtBeginDocument{%
  \providecommand\BibTeX{{%
    \normalfont B\kern-0.5em{\scshape i\kern-0.25em b}\kern-0.8em\TeX}}}

\setcopyright{acmcopyright}
\copyrightyear{2018}
\acmYear{2018}
\acmDOI{10.1145/1122445.1122456}

\acmConference[Woodstock '18]{Woodstock '18: ACM Symposium on Neural
  Gaze Detection}{June 03--05, 2018}{Woodstock, NY}
\acmBooktitle{Woodstock '18: ACM Symposium on Neural Gaze Detection,
  June 03--05, 2018, Woodstock, NY}
\acmPrice{15.00}
\acmISBN{978-1-4503-XXXX-X/18/06}

\copyrightyear{2020} \acmYear{2020} \setcopyright{acmcopyright}\acmConference[MM '20]{Proceedings of the 28th ACM International Conference on Multimedia}{October 12--16, 2020}{Seattle, WA, USA} \acmBooktitle{Proceedings of the 28th ACM International Conference on Multimedia (MM '20), October 12--16, 2020, Seattle, WA, USA} \acmPrice{15.00} \acmDOI{10.1145/3394171.3413621} \acmISBN{978-1-4503-7988-5/20/10}

\begin{document}
\fancyhead{}

\title{Learning Deep Multimodal Feature Representation\\with Asymmetric Multi-layer Fusion}

\author{Yikai Wang$^{1*}$, Fuchun Sun$^1$, Ming Lu$^2$, Anbang Yao$^2$} 
\affiliation{ 
      $^1$Beijing National Research Center for Information Science and Technology$\,$(BNRist),\\ State Key Lab on Intelligent Technology and Systems,\\ Department of Computer Science and Technology, Tsinghua University\\
      $^2$Cognitive Computing Laboratory, Intel Labs China
    }
\email{{wangyk17@mails.,fcsun@}tsinghua.edu.cn, {ming1.lu,anbang.yao}@intel.com}

%
\renewcommand{\shortauthors}{Y. Wang, F. Sun, M. Lu and A. Yao}


\begin{abstract}
We propose a compact and effective framework to fuse multimodal features at multiple layers in a single network. The framework consists of two innovative fusion schemes. Firstly, unlike existing multimodal methods that necessitate individual encoders for different modalities, we verify that multimodal features can be learnt within a shared single network by merely maintaining modality-specific batch normalization layers in the encoder, which also enables implicit fusion via joint feature representation learning. Secondly, we propose a bidirectional multi-layer fusion scheme, where multimodal features can be exploited progressively. To take advantage of such scheme, we introduce two asymmetric fusion operations including channel shuffle and pixel shift, which learn different fused features with respect to different fusion directions. These two operations are parameter-free and strengthen the multimodal feature interactions across channels as well as enhance the spatial feature discrimination within channels. We conduct extensive experiments on semantic segmentation and image translation tasks, based on three publicly available datasets covering diverse modalities. Results indicate that our proposed framework is general, compact and is superior to state-of-the-art fusion frameworks.
\end{abstract}



\maketitle

\renewcommand*{\thefootnote}{\fnsymbol{footnote}}
\footnotetext[0]{$^*\,$This work was done when Yikai Wang was an intern at Intel Labs China, supervised by Anbang Yao who is responsible for correspondence.}
\renewcommand*{\thefootnote}{\arabic{footnote}}

\section{Introduction}
\label{intro}

Understanding complex visual scenes is an essential prerequisite for robots and autonomous vehicles to operate in real-world. With the increasing availability of multimodal sensors, fusing information collected by these sensors has shown improved performance on several tasks, like scene recognition, semantic segmentation, etc. In typical multimodal settings, RGB and depth inputs have been widely used to date. Besides, other multimodal inputs such as  normals, shadings, textures and edges, are also discussed to some extent. It has been largely verified that the way to effectively fuse multimodal features is essential to the final performance.  In this work, we mainly focus on multimodal inputs which are aligned in pixel-level (e.g., RGB, depth and shade), and tackle two drawbacks of existing multimodal fusion works relying on deep neural networks, by proposing two innovative fusion schemes.

Firstly, existing multimodal training methods follow a common design practice that an individual encoder branch is specialized for each modality. For example, regarding to the semantic segmentation task, FuseNet \cite{DBLP:conf/accv/HazirbasMDC16}, RDFNet \cite{DBLP:conf/iccv/LeePH17}, SSMA \cite{DBLP:journals/ijcv/RussakovskyDSKS15} adopt two equal-sized encoders for RGB and depth inputs respectively. The underlying reason may be that for different modalities, different characteristics and feature statistics are not compatible in a single model. However, despite of the heavy parameter load, it prevents the possibility of multimodal features to implicitly fuse during joint training. To tackle this issue, we find that individual encoders are not necessary for multimodal inputs as long as Batch Normalization layers (BNs) \cite{DBLP:conf/icml/IoffeS15} are privatized. Specifically, we share all convolutional filters in both encoder and decoder, but adopt modality-specific BNs in the encoder. Modality-specific BNs estimate the channel-wise running mean and variance of activations for each modality separately, and also learn individual channel-wise scale and bias. We empirically verify the effectiveness of this scheme on various modalities, including RGB, depth, normal, shade, etc. On the one hand, this training scheme largely reduces the number of parameters needed for multimodal training. On the other hand, it allows a single network to exploit multimodal features simultaneously, which improves the generalization of convolutional neural networks and achieves better training performance in practice.

Secondly, key ingredients of multimodal fusion include how to design fusion blocks and where to implement fusion. It is verified that exploiting multi-layer fusion, i.e., fusing multimodal features at multiple stages of the network, will improve the fusion performance. In early multimodal fusion works, fusion could be simply realized by feature concatenation, addition or average.  Recently, to enable more powerful feature alignment, some multi-layer fusion works adopt a pile of $3\times3$ convolutional layers \cite{DBLP:conf/iccv/LeePH17} or attention-based designs \cite{DBLP:journals/ijcv/RussakovskyDSKS15} for fusion. 
However, as we will explain, these fusion methods tend to learn symmetric features when followed by a pointwise convolutional layer. This can be simply understood as A$\to$B and B$\to$A fusion leading to the same expressive ability of feature maps (regardless of the order of channels). In this work, we propose a bidirectional fusion scheme, enabling more sufficient multimodal feature fusion. We argue that although existing symmetric fusion methods are suitable for the unidirectional fusion, they are not very compatible with the bidirectional fusion. Besides, along with the emergence of powerful yet complex fusion blocks, increasing amounts of parameters are introduced when fusing multimodal features at multiple layers. To fit the bidirectional fusion scheme, we propose two brand-new asymmetric multimodal fusion operations. Being a fusion in the cross-channel direction, the channel shuffle operation strengthens the multimodal feature interactions across channels, improving the holistic feature representation ability. Being a fusion in the spatial direction within each channel, the pixel shift operation tends to enhance spatial feature discrimination, capturing fine-grained info at object edges, especially for small and thin objects. Both channel shuffle and pixel shift are plug-in and parameter-free operations. 

We apply our schemes to two tasks including semantic segmentation and image translation. Our work is verified on three different datasets containing diverse application scenarios ranging from urban city driving scenes to indoor scenes, covering rich modalities  including RGB, depth, shade, normal, texture, and edge. Different network architectures containing ResNet \cite{DBLP:conf/cvpr/HeZRS16}, Xception65 \cite{DBLP:conf/cvpr/Chollet17} and U-Net \cite{DBLP:conf/miccai/RonnebergerFB15} are adopted as backbones.  Experimental results prove the effectiveness and generalization of our proposed schemes. 

Main contributions of this work can be summarized as follows:
\leftmargini=7mm
\begin{itemize}
\setlength\itemsep{0em}

\item We verify that multimodal inputs can be fed into a single network with shared parameters and individual BNs for each modality in the encoder, achieving even higher performance than the common practice which uses individual networks.

\item We propose two asymmetric parameter-free fusion operations, enabling bidirectional multi-layer fusion from both channel-level and pixel-level perspectives. These operations strengthen the multimodal feature interactions across channels as well as enhance the spatial feature discrimination.

\item By merely introducing about 0.1\% additional parameters on a given unimodal network, our fusion method is able to outperform state-of-the-art fusion methods on several datasets.

\end{itemize}

\section{Related Works}
\label{related}
\textbf{Multimodal Fusion.}
Methods to exploit multimodal information have been studied for decades \cite{DBLP:conf/eccv/SilbermanHKF12,DBLP:conf/cvpr/GuptaAM13,DBLP:conf/eccv/GuptaGAM14,DBLP:conf/accv/HazirbasMDC16,DBLP:journals/tcyb/LinZJLH20}, which allow better understanding of visual scenes compared to learning with unimodal inputs. Prior works usually rely on hand engineered or learned features extracted from each individual modality and combine features together with designed fusion structures. In \cite{DBLP:conf/eccv/SilbermanHKF12}, Markov random fields are explored for indoor segmentation based on RGB and depth inputs. \cite{DBLP:conf/cvpr/GuptaAM13} improves RGB-D recognition performance by making use of the  constructed geometric contour from depth data. More recently, with the success of deep convolutional neural networks, a series of multimodal fusion schemes are proposed for end-to-end feature fusion. Regarding to the fusion position, these works can be categorized into single-layer fusion and multi-layer fusion methods. In schemes of single-layer fusion,  multimodal features are usually merged into one branch at a particular layer. For example, \cite{DBLP:journals/corr/abs-1301-3572,DBLP:conf/iccv/EigenF15} stack multimodal inputs by channel-wise concatenation and then feed them to the network. \cite{DBLP:conf/eccv/WangWTSW16} designs a transformation layer which fuses multimodal features between the encoder and decoder. \cite{DBLP:journals/corr/LongSD14,DBLP:conf/cvpr/ChengCLZH17} explore multimodal feature fusion at the prediction side. However, it has been verified that single-layer fusion methods  can not effectively exploit multimodal features, especially for addressing high-resolution predictions \cite{DBLP:conf/iccv/LeePH17,DBLP:journals/ijcv/RussakovskyDSKS15}. Besides, in \cite{DBLP:conf/cvpr/ZengTHYSCW19}, it shows that single-layer fusion is sensitive to the noises in multimodal data. Owing to these factors, multi-layer fusion methods become popular, which combine multimodal features at multiple levels, usually at every downsampling stage of a network. Existing multimodal multi-layer  fusion schemes can be further classified into two kinds. The first kind is to directly send the fused features to the decoder side. RDFNet \cite{DBLP:conf/iccv/LeePH17} adopts multi-layer fusion at four downsampling stages of the ResNet (encoder), iteratively refines the fused features with the similar idea in RefineNet \cite{DBLP:conf/cvpr/LinMSR17}, and then sends these fused features to the decoder. SSMA \cite{DBLP:journals/ijcv/RussakovskyDSKS15} fuses multimodal features at mid-level and high-level with an attention-based mechanism for feature calibration, and as well sends the fused features to the decoder side. However, this kind of fusion methods prevents the encoder to exploit multimodal features.
The second kind is to apply fused features to one of the branches in the encoder for in-depth feature exploiting. Typical works can be traced back to FuseNet \cite{DBLP:conf/accv/HazirbasMDC16}, which trains two individual branches to learn RGB and depth features, where multiple skip connections in the encoder from the depth branch to RGB branch are used for fusion. \cite{DBLP:conf/cvpr/ZengTHYSCW19} also adopts two branches for learning RGB-D features respectively, and merges depth features into RGB branch at all downsampling layers in a hierarchical manner.
This kind of methods is illustrated in Figure \ref{fusion} (a), which will be also discussed later. Such scheme is unidirectional and straightforward, still lacking rich feature interactions and thus may be not sufficient for fusion. To tackle this drawback, we design a bidirectional fusion scheme to improve fusion performance.

\textbf{Shuffle and Shift.} In group convolutions \cite{DBLP:journals/corr/HowardZCKWWAA17,DBLP:conf/cvpr/Chollet17},  outputs that correspond to each channel are only related with a portion of input channels.  To address this issue, channel shuffle is presented in ShuffleNet \cite{DBLP:conf/cvpr/ZhangZLS18}  to enhance the information flow across different input channel groups, so that each channel is correlated with all groups. Inspired by ShuffleNet, a recent work \cite{DBLP:journals/corr/abs-1911-11319} allocates each frame feature into groups and then aggregates the grouped features via temporal shuffle operation.  To date, the proposed shuffle operations are mostly adopted for strengthening correlations among different groups. In this work, we propose to use the channel shuffle operation for multimodal fusion, partially aiming to promote feature interaction among multimodal features, but also to make use of its asymmetric property, which will be analyzed in Section \ref{sec:shuffleshift}. There are also some research works that apply shift operations, which are related to our design. Early in \cite{DBLP:conf/cvpr/WuWYJZGGGK18}, stacking pixel shift layers is treated as parameter-free and FLOP-free operations to improve spatial information communication. The shift design is further improved in \cite{DBLP:conf/cvpr/ChenXZP19}, where only a few shift operations are needed, instructed by shift operation penalty and quantization-aware shift learning method. Also in \cite{DBLP:conf/iccv/LinGH19}, a temporal shift module is proposed to shift a portion of channels along the temporal dimension, aiming to boost information exchange among neighboring frames. In this work, we extend the idea of pixel shift to facilitate spatial correlations among multimodal features, and again, we point out that the shift operation can be another asymmetric fusion operation which is compatible with the proposed bidirectional fusion scheme.

\section{Approach}
In this section, we propose two multimodal fusion schemes. The first is a parameter-sharing scheme which compresses the model size and enables implicit feature fusion. The second is a bidirectional fusion scheme which enables explicit and sufficient feature fusion.
\subsection{Parameter-sharing Scheme}
\label{sec:parameter_sharing}
\begin{figure}[t]
\centering
\includegraphics[scale=0.2]{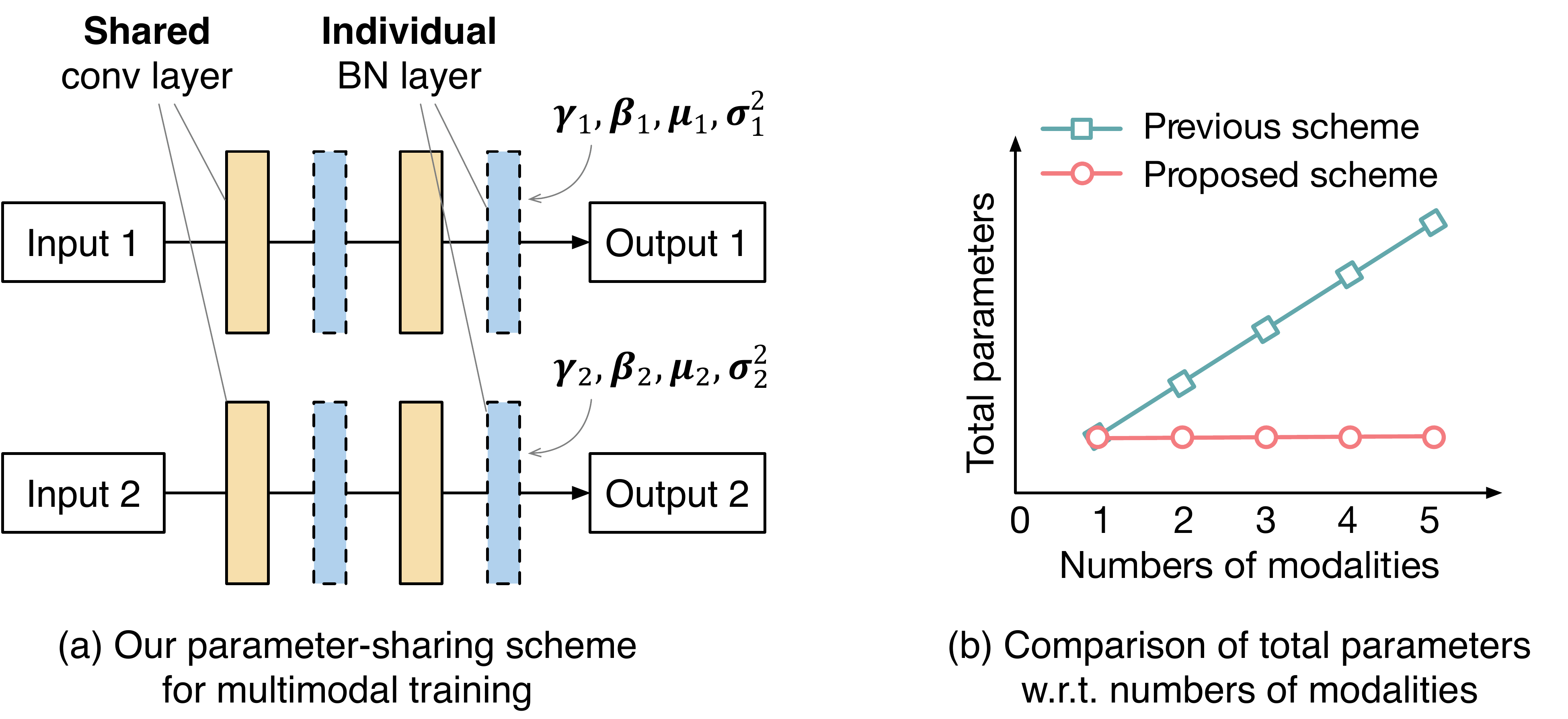}
\vskip-0.2em
\caption{\textmd{(a) A compact multimodal fusion scheme, with shared parameters for convolutional layers (also for fully-connected layers, if any) and individual BN parameters. (b) A comparison of total parameters for feature encoding between existing multimodal fusion schemes and ours. With the increasing of total modalities, the size of our scheme is nearly unchanged.}}
\label{pics:parameter}
\end{figure}

As illustrated in Figure \ref{pics:parameter}(a), we provide a multimodal fusion scheme with two input modalities as an example. Unlike existing works which necessitate individual encoders for multiple modalities, we propose that by sharing convolutional parameters but leaving Batch Normalization layers (BNs) \cite{DBLP:conf/icml/IoffeS15} modality-specific, we are able to train a single network for multiple modalities. 
In modern deep neural networks, the mechanism of using BNs has become one of the most successful architectural innovations. A BN layer whitens activations over a mini-batch of features, and transforms the whitened results with channel-wise affine parameters, including scale  and bias  which provide the possibility of linearly transforming whitened activations to any scales. Sharing network parameters but privatizing BNs has been proved to be effective for efficient model adaption when considering multiple tasks or multiple domains \cite{DBLP:conf/iclr/LiWS0H17,DBLP:conf/iclr/MudrakartaSZH19,DBLP:conf/iclr/YuYXYH19,wang2020rsnets}. Inspired by this, we extends the idea of privatizing BNs for multimodal training, where activation statistics of different modalities are normalized separately, and channel-wise scale and bias of BNs are also learned individually for each modality. Network parameters apart from BNs are shared for all modalities. Specifically, assuming there are $S$ modalities for fusion, for the $s^{\text{th}}$ modality, where $s\in\{1,2,\cdots,S\}$, privatizing BN can be formulated as:
\begin{equation}\label{norm}
\bm{y}_s=\bm{\gamma}_s\cdot \frac{\bm{x}_s-\bm{\mu}_s}{\sqrt{\bm{\sigma}_s^2+\epsilon}}+\bm{\beta}_s,
\end{equation}
where $\bm{x}_s,\bm{y}_s\in\mathbb{R}^{N\times C\times H \times W}$ are the input and output feature maps of the $s^{\text{th}}$ modality respectively; $N,C,H,W$ denote the batch size, the number of channels, the height and width of the feature map respectively; $\bm{\mu}_s,\bm{\sigma}_s^2\in\mathbb{R}^{C}$ are the mean and standard deviation values of input activations over the current mini-batch of the $s^{\text{th}}$ modality, calculated by $\bm{\mu}_s=\frac{1}{NHW}\sum_{n,h,w}\bm{x}_s$ and $\bm{\sigma}_s^2=\frac{1}{NHW}\sum_{n,h,w}(\bm{x}_s-\bm{\mu}_s)^2$; besides, $\bm{\gamma}_s,\bm{\beta}_s$ $\in\mathbb{R}^{C}$ are learnable scale and bias with respect to the $s^{\text{th}}$ modality; $\epsilon$ is a small constant to avoid division by zero.

\begin{figure}[t]
\centering
\includegraphics[scale=0.2]{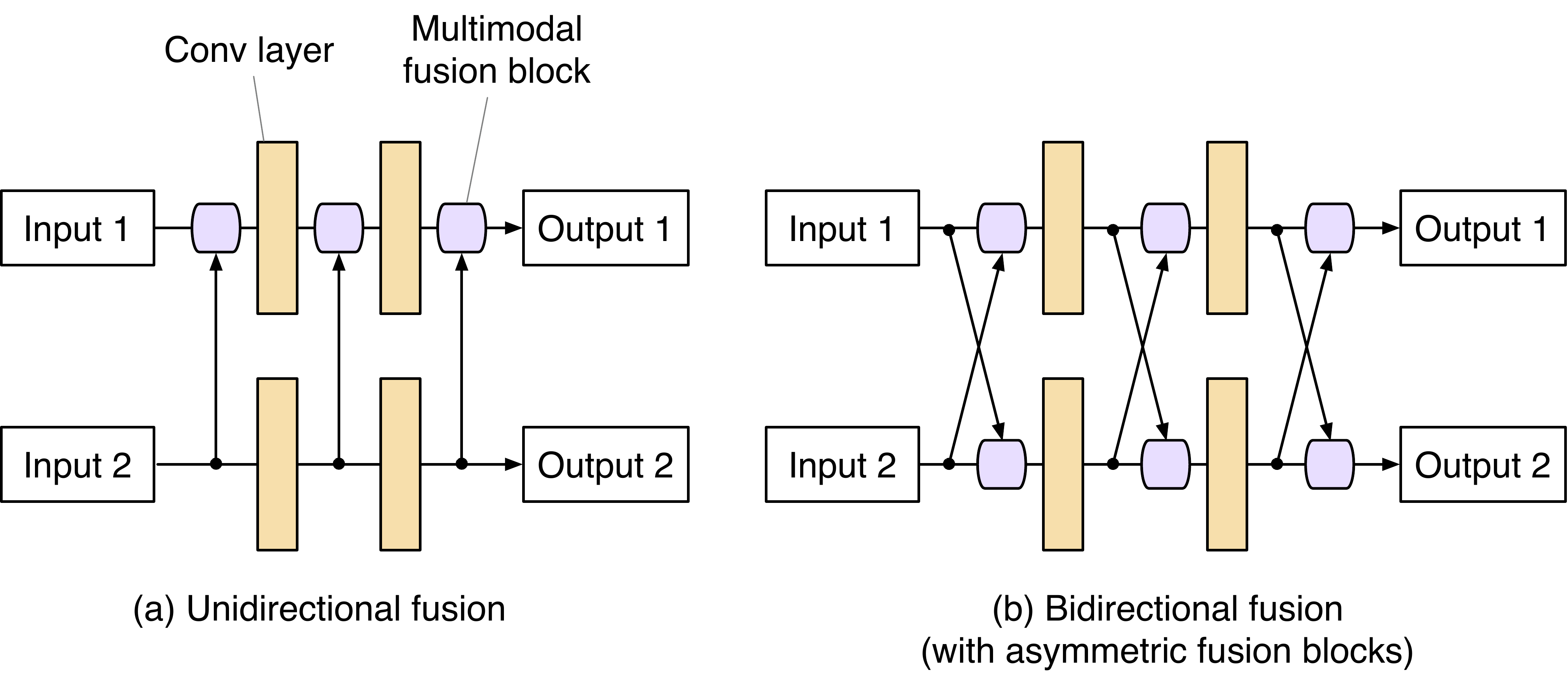}
\vskip-0.3em
\caption{\textmd{(a) Existing multimodal multi-layer fusion schemes mostly adopt unidirectional fusion. (b) Our proposed bidirectional fusion scheme enables each branch to exploit multimodal features. In order to take full advantage of this scheme, we need to design new asymmetric fusion methods.}}
\label{fusion}
\vskip-0.5em
\end{figure}

Note that the parameter-sharing indicates sharing all convolutional filters in both encoder and decoder, but privatizing BNs indicates using modality-specific BNs merely in the encoder. We find sharing BNs in the decoder part achieves better results especially for the multimodal image translation task.

This training scheme largely compresses model parameters for multimodal fusion. Besides that, learning  with shared parameters facilitates interior interactions of multimodal features, which even brings performance improvements. Verification results will be discussed in Table \ref{tabs:sharing}, where nearly 50\% parameters reduction is obtained with even higher evaluation scores compared with using individual encoders. Such paramter-sharing scheme is not limited to two modalities, which will be verified in Table \ref{tabs:imgtranslation}.

\begin{figure*}[t]
\centering
\includegraphics[scale=0.4]{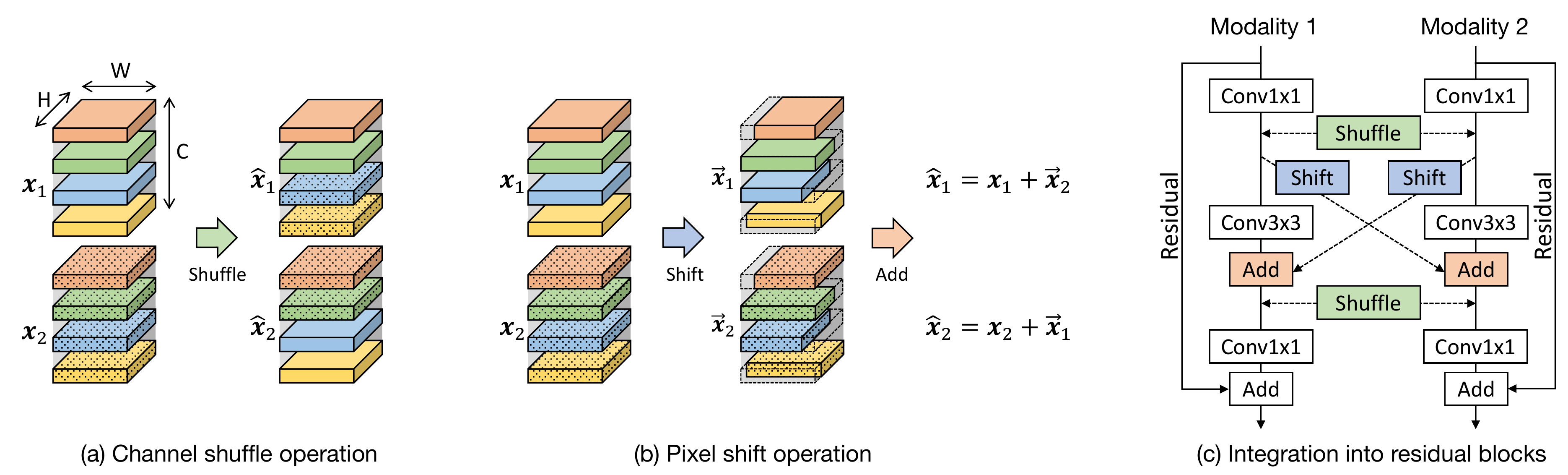}
\caption{\textmd{Proposed two asymmetric multimodal fusion operations that are compatible with bidirectional fusion scheme. In the figure, we use a hat sign to represent the fused feature, i.e., $\widehat{\bm{x}}_1=\mathcal{F}(\bm{x}_1, \bm{x}_2),\,\widehat{\bm{x}}_2=\mathcal{F}(\bm{x}_2, \bm{x}_1)$. Best be viewed in color.  (a) Channel shuffle operation exchanges a portion of features with respect to the same-indexed channels. (b) Pixel shift operation performs pixel-wise shifts on four directions, and uses the addition across multimodal features for fusion. (c) We integrate both operations into residual blocks of ResNet architectures for illustration. Blocks in color and dashed lines indicate our inserted structures. Note that the colors of blocks are consistent with the colors of arrows in the first two subfigures, for better understanding.}}
\label{shuffle_shift}
\end{figure*}

\subsection{Bidirectional Fusion Scheme}
\label{sec:shuffleshift}
As described in Section \ref{related}, early works usually fuse multimodal features at one particular layer, while it has been recently verified that multi-layer fusion (fuse at multiple layers) can exploit supplementary information more adequately. Regarding typical multi-layer fusion works \cite{DBLP:conf/accv/HazirbasMDC16,DBLP:conf/cvpr/ZengTHYSCW19}, multimodal features are usually fused into one branch and are then further exploited by the encoder. The scheme can be shown as Figure \ref{fusion}(a), where features learned from the second branch are merged into the first branch. This scheme allows the first branch of encoder to exploit fused features. However, as the fusion is unidirectional, the second branch remains unimodal from beginning to end and thus cannot bring informative features at later fusion layers. Besides, two branches would be highly unbalanced during training, and such unbalance may impact the fusion performance. In our experiments, we find for the unidirectional fusion scheme, different fusion directions lead to very different performance.

From this point of view, it would be worthwhile to design a new kind of multi-layer fusion that can overcome the aforementioned drawback, and improve the fusion performance. To this end, we propose a bidirectional fusion scheme for multi-layer  fusion, illustrated in Figure \ref{fusion}(b). In this scheme, multimodal features of different encoder branches are merged mutually, enabling rich feature interactions. However, we find commonly used fusion operations, such as concatenation and average, are not very compatible with the bidirectional fusion scheme. To make it clear, we provide a following definition using two modalities as an example.

\textbf{Definition.} Let $\mathcal{F}(\cdot,\cdot;\bm{\theta})$ be a fused feature of two modalities, where $\bm{\theta}$ denotes the internal parameters of the fusion block. Let $\mathcal{C}$ be a single pointwise convolutional layer. We define a fusion block as \textbf{symmetric} if for any $\bm{\theta}_1,\mathcal{C}_1$, there exist $\bm{\theta}_2,\mathcal{C}_2$ s.t. $\mathcal{C}_1(\mathcal{F}(\bm{x}_1, \bm{x}_2;\\\bm{\theta}_1))=\mathcal{C}_2(\mathcal{F}(\bm{x}_2, \bm{x}_1;\bm{\theta}_2))$, holding for any two feature maps $\bm{x}_1, \bm{x}_2$. We define a fusion block as \textbf{asymmetric} if the definition of symmetric fusion does not hold.

The reason for introducing a convolutional layer $\mathcal{C}$ into the definition is because in practice, most fused features are followed by a convolutional layer which further mixes features along the channel. Regarding common fusion methods, it is obvious that addition and average operations are symmetric. The concatenation operation can be proved symmetric when exchanging the order of the same-length outputs for $\bm{x}_1$ and $\bm{x}_2$, which can be realized by their following $\mathcal{C}_1$ and $\mathcal{C}_2$. For these three fusion methods, internal fusion parameters $\bm{\theta}=\emptyset$. Recently proposed fusion methods that apply internal convolutional layers, i.e., $\bm{\theta}\ne\emptyset$, for example attention-based fusion blocks in SSMA \cite{DBLP:journals/ijcv/RussakovskyDSKS15}, and MMF blocks in RDFNet \cite{DBLP:conf/iccv/LeePH17}, can also be proved symmetric when $\bm{\theta}_2$ is obtained by exchanging two groups of modality-specific parameters of $\bm{\theta}_1$. These commonly used symmetric fusion methods are not very compatible with the bidirectional fusion scheme. See Figure \ref{fusion}(b), during multimodal training, as both final outputs are usually applied with the same supervision signals, features fused by symmetric fusion methods at both branches tend to learn similar representations (potentially can be the same). This would bring redundant information at both encoder branches.

Based on the above analysis, we propose two kinds of asymmetric fusion operations to fit the bidirectional fusion scheme, called \textbf{AsymFusion}. These operations include channel shuffle and pixel shift, which are both parameter-free. We first introduce the designs of both operations, and show how to integrate them into popular network architectures. 

\textbf{Channel Shuffle.} To strengthen the interaction of multimodal information flow across channels, we propose the channel shuffle operation. As illustrated in Figure \ref{shuffle_shift}(a), when given two feature maps $\bm{x}_1,\bm{x}_2\in\mathbb{R}^{C\times H\times W}$, channel shuffle fuses two features by exchanging features corresponding to a portion of channels. Given a channel spilt point $T$, satisfying $1<T<C,T\in\mathbb{Z}$, the channel shuffle operation can be described as:
\begin{equation}
\begin{split}
&\mathcal{F}(\bm{x}_1, \bm{x}_2)=\bm{x}_1[1,\cdots,T]\;||\;\bm{x}_2[T+1,\cdots,C],\\
&\mathcal{F}(\bm{x}_2, \bm{x}_1)=\bm{x}_2[1,\cdots,T]\;||\;\bm{x}_1[T+1,\cdots,C],
\end{split}
\end{equation}
where $||$ indicates channel-wise concatenation; $1,\cdots,T$ and $T+1,\cdots,C$ indicate channel indices, which in our experiments, make up 70\%, 30\% channels respectively. According to this formulation, no additional parameters are introduced, i.e., $\bm{\theta}=\emptyset$.

An essential property of the shuffle operation leading to its asymmetry is that, two features after shuffle have no feature overlaps. Hence, it is straightforward to find input feature maps $\bm{x}_1, \bm{x}_2$ and $\mathcal{C}_1$, such that no $\mathcal{C}_2$ is able to result in the same features, which means fusion by channel shuffle is asymmetric. 

\textbf{Pixel Shift.}\label{sec:pixel_shift} 
To improve spatial information communication of multimodal features, we introduce the second asymmetric fusion operation which uses pixel shift on feature maps, as shown in Figure \ref{shuffle_shift}(b). Fusion by pixel shift contains two steps. Supposing the number of channels $C$ is divisible by $4$. For the first step, we divide the feature into four groups, each having $C/4$ channels, and shift one pixel for every group with one of four spatial directions. This operation can be formulated as:
\begin{equation}
\begin{split}
&\vec{\bm{x}}_1[c,h,w]=\mathcal{O}(\bm{x}_1)[c,h+\alpha_c+1,w+\beta_c+1],\\
&\vec{\bm{x}}_2[c,h,w]=\mathcal{O}(\bm{x}_2)[c,h+\alpha_c+1,w+\beta_c+1],
\end{split}
\end{equation}
where $\mathcal{O}(\cdot)$ indicates a zero padding; $\alpha_c,\beta_c$ are position indicators, $\alpha_c=[0,-1,0,1]_{\lfloor c/4\rfloor},\;\beta_c=[-1,0,1,0]_{\lfloor c/4\rfloor}$; and adding one pixel on position indicators is due to the zero padding.

For the second step, each fused feature is obtained by adding the original feature and the shifted feature of the other modality:
\begin{equation}
\begin{split}
&\mathcal{F}(\bm{x}_1, \bm{x}_2)=\bm{x}_1+\vec{\bm{x}}_2,\\
&\mathcal{F}(\bm{x}_2, \bm{x}_1)=\bm{x}_2+\vec{\bm{x}}_1,
\end{split}
\end{equation}
again, the shift operation brings no additional parameters, i.e., $\bm{\theta}=\emptyset$. To illustrate that the shift operation is asymmetric, we simply let $\bm{x}_1=0,\bm{x}_2\ne0$. For any $\mathcal{C}_1$, if the definition of symmetric fusion is satisfied, there should always exist $\mathcal{C}_2$ such that $\mathcal{C}_1(\vec{\bm{x}}_2)=\mathcal{C}_2(\bm{x}_2)$. However, this statement cannot be true, as a shifted feature no longer keeps its original pixel alignments across channels.

Both fusion operations introduce no additional parameters, and are also FLOP-efficient. They not only have advantages on asymmetry, but also introduce channel-wise and spatial-wise feature interactions across modalities, respectively, making them more effective for multimodal multi-layer fusion. In Section \ref{discuss}, we will show that adopting shift operations in a network may improve predictions for fine-grained objects,  and strengthen the representation ability to discriminate rich edge-aware information from context.

Since we have presented two asymmetric fusion operations that can be compatible with the bidirectional fusion scheme, we now consider integrating them into convolutional networks. We adopt residual blocks of ResNet \cite{DBLP:conf/cvpr/HeZRS16} as an example. As shown in Figure \ref{shuffle_shift}(c), we insert both fusion operations into the residual blocks of ResNet. As the first Conv$1\times1$ layer in each residual block performs compression on channel dimension, we speculate that adding shuffle and shift between the two Conv$1\times1$ layers where features have less channels, will lead to lower information loss during fusion. Specifically, we choose to apply two shuffle operations, for a more sufficient channel fusion, after the first Conv$1\times1$ layer and before the last Conv$1\times1$ layer respectively. Besides, we insert the shift operations after the first shuffle. We empirically find that adding the shifted features to features after the Conv$3\times3$ layers will lead to better performance. Although the pixel shift operation constantly shifts one pixel on a feature map, the corresponding shift of the original image will vary according to the feature map size. To mix different extents of side effects caused by pixel shifts, we apply our fusion approaches at every downsampling stage of the network. Besides fusing at downsampling stages, we also try applying shuffle and shift to other layers but do not observe further improvements. 

\section{Experiments}
\label{experiments}

In order to verify the generalization of our proposed schemes, we conduct experiments on two tasks, including semantic segmentation and image translation. The benchmark is performed on three datasets with diverse environments ranging from urban city scenes to indoor scenes, covering a variety of different modalities, including RGB, depth, shade, normal, texture, and edge. Besides, we apply the proposed fusion structures on different network architectures, such as ResNet, Xception65 and U-Net.

\subsection{Datasets}
We consider  two indoor datasets NYUDv2 \cite{DBLP:conf/eccv/SilbermanHKF12}, Taskonomy \cite{DBLP:conf/cvpr/ZamirSSGMS18}, and an outdoor dataset CityScapes \cite{DBLP:conf/cvpr/CordtsORREBFRS16}.

\textbf{NYUDv2} is one of the most popular RGB-D datasets in the literature for  indoor scene labeling, containing 1449 densely labeled pairs of RGB and depth frames. Following the standard settings, we use 795 training RGB-D pairs and 654 testing RGB-D pairs, and we evaluate our network on 40 classes with labels provided by \cite{DBLP:conf/cvpr/GuptaAM13}.

\textbf{Cityscapes} is a RGB-D dataset for street scene understanding which becomes one of the standard benchmarks. The dataset contains images from 50 different cities, during varying seasons, lighting and weather conditions. The dataset provides 2875 images for training and 500 images for validation. We \textbf{do not} use the supplementary training data with coarse annotations.

\textbf{Taskonomy} is a large-scale dataset for indoor scene understanding, where each image has other corresponding modalities such as depth, normal, shade, texture, etc., with detailed annotations. We adopt its official sub-dataset containing 9000 samples. As there are no public papers or benchmarks that provide train-test splits to now, we randomly divide the data into 8000 samples for training and the rest 1000 samples for testing, and guarantee no scene overlaps in training and testing samples.

\subsection{Implementation Details}
\label{sec:imple}
For semantic segmentation tasks, in order to facilitate comparison with previous fusion approaches, we choose RefineNet \cite{lin2019refinenet} for the dataset NYUDv2 and DeepLabv3+ \cite{DBLP:conf/eccv/ChenZPSA18} for the dataset Cityscapes, with backbone architectures ResNet \cite{DBLP:conf/cvpr/HeZRS16} and Xception65 \cite{DBLP:conf/cvpr/Chollet17} respectively. For ResNet architectures, we implement the fusion blocks at all downsampling stages. To make the structure clear, we provide a table showing structure details in supplementary materials. For Xception65, we apply our fusion designs to the last blocks of the entry flow, middle flow and exit flow respectively. As a standard evaluation method, We apply test-time multi-scale evaluation for all experiments and obtain final predications as average results.

\begin{figure*}[t]
\centering
\includegraphics[scale=0.31]{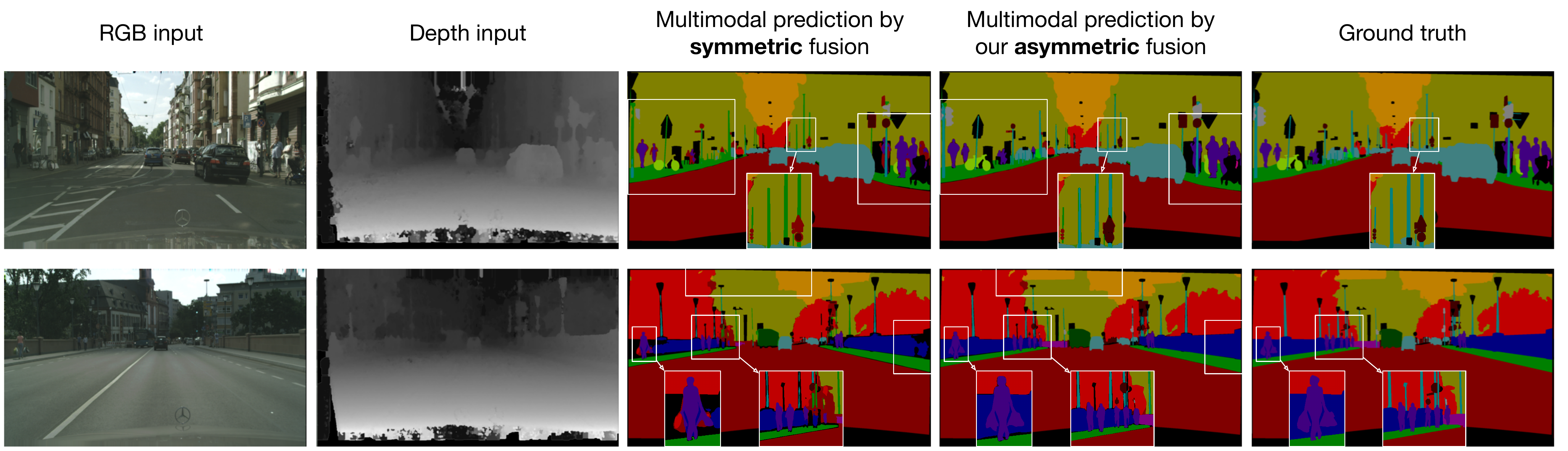}
\vskip-0.5em
\caption{\textmd{Illustrative results of semantic segmentation on Cityscapes dataset. In the third column, we choose the attention-based fusion method as the baseline, which achieves state-of-the-art performance as shown in Table  \ref{tabs:city}. We provide results predicted by our asymmetric fusion method in the fourth column. Regions with sharp predication differences are indicated with white frames. \textbf{Best viewed in color at zoom 300\%.}}}
\label{pic:city}
\end{figure*}

For the image translation task, following pix2pix, we adopt a U-Net \cite{DBLP:conf/miccai/RonnebergerFB15} generator, with an eight-layer encoder and an eight-layer decoder. We adopt multi-layer fusion and apply the fusion with shuffle and shift at the $3^{\text{rd}}$, $5^{\text{th}}$, $7^{\text{th}}$ layers of the encoder respectively. We use a discriminator with five convolutional layers for downsampling. 

Parameter sharing strategies have been described in Section \ref{sec:parameter_sharing}. Specifically, for both semantic segmentation and image translation tasks, we share all convolutional parameters and privatize BNs for different modalities in the encoder; we directly share all parameters including BNs in the decoder.

For the semantic segmentation task, we learn an ensemble at the final predictions to further fuse multimodal prediction scores. For $S$ modalities, the ensemble is learned using a group of importance scores $\bm{\alpha}\!=[\alpha_1\,\alpha_2\,\cdots\,\alpha_S]$, satisfying $\bm{\alpha}\!\ge\!0$, $\sum_{s=1}^S\alpha_s\!=\!1$, which can be easily implemented with a softmax function. We then adopt the technique of knowledge distillation to force predications of different modalities to mimic the learned ensemble prediction. We find this strategy brings additional improvements for multimodal fusion with negligible extra costs (only $S$ extra parameters).

More implementation details, including learning rate and epoch settings, and the method to extend our bidirectional fusion scheme to more modalities, are provided in supplementary materials.

\subsection{Results}
\textbf{Semantic Segmentation.} We report results on both NYUDv2 and Cityscapes datasets in Table \ref{tabs:nyudv2} and Table \ref{tabs:city} respectively. Besides commonly used metrics for semantic segmentation, including pixel accuracy, mean accuracy, and intersection over union (IoU), we also compare total parameters of models. In Table \ref{tabs:nyudv2}, we compare our AsymFusion with four state-of-the-art methods. For a quick comparison with RDFNet, which is a RGB-D fusion scheme using RefineNet as its decoder. Using the same encoder ResNet101, total parameters of RDFNet (366.7M) are much larger than RefineNet (118.1M), making it unfriendly for model deployment. In sharp contrast, when given RefineNet (ResNet101) as the unimodal architecture, our fusion model only has 118.2M total parameters, with less than 0.1M additional parameters than RefineNet, and less than $1/3$ parameters compared with RDFNet. Similarly, using ResNet152 as encoder, our model merely introduces 0.11\% additional parameters compared with RefineNet. Under this extremely compact setting, our fusion models still outperform other methods with a large margin (over 1\% absolute gain for IoU). In Table \ref{tabs:city}, we compare our fusion method to state-of-the-art methods on Cityscapes validation dataset, and our fusion method outperforms previous methods. We do not use the supplementary training set including 20,000 coarse annotations, considering the training costs. Illustrative results shown in Figure \ref{pic:city} verify that our model captures fine-grained details.

\begin{figure*}[t]
\centering
\includegraphics[scale=0.22]{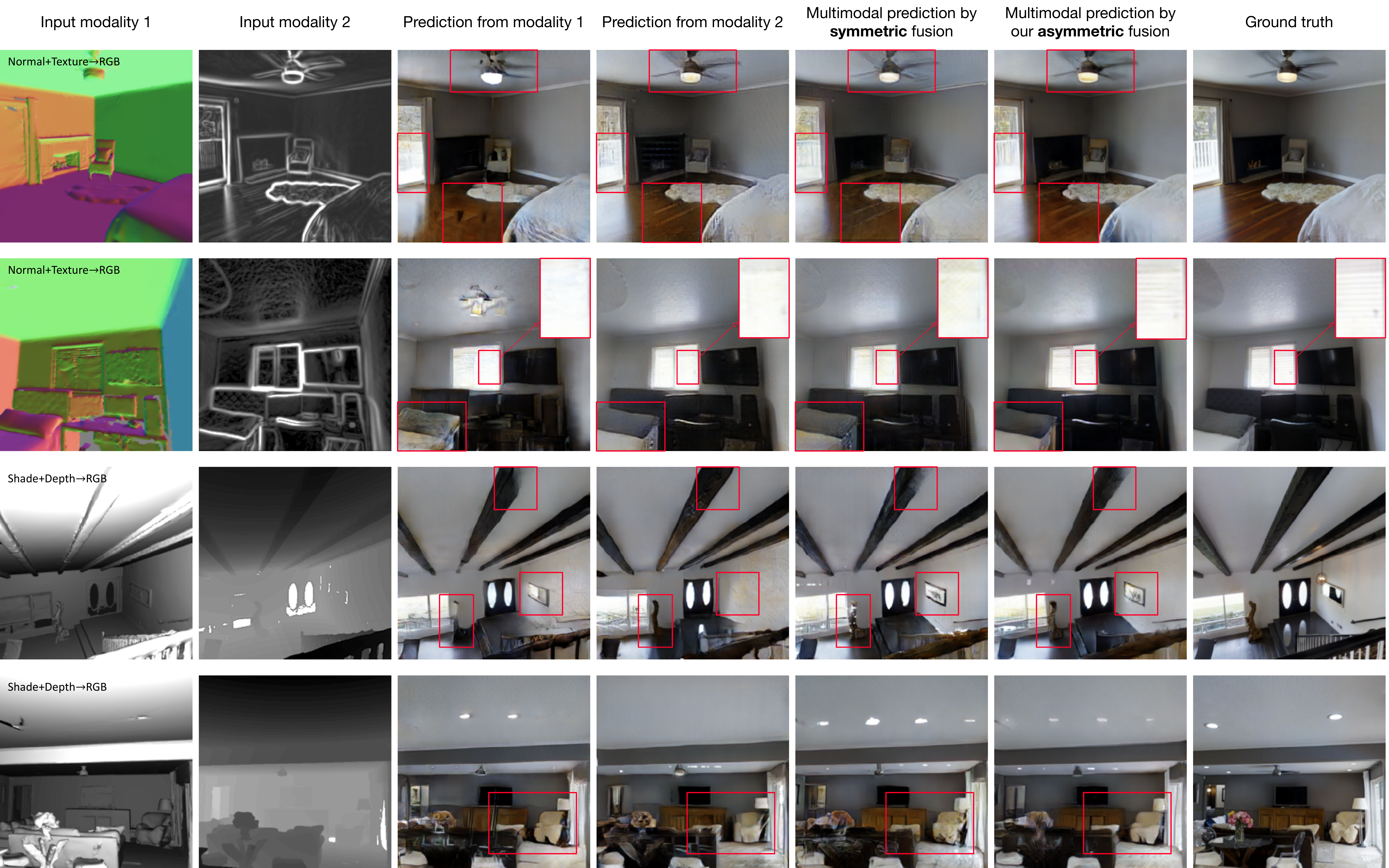}
\vskip-0.5em
\caption{\textmd{Image translation with two modalities as inputs, tested on Taskonomy dataset. We provide the result predicted by each single modality respectively, and compare the multimodal fusion performance using symmetric fusion and our asymmetric fusion. For symmetric fusion method, we adopt attention-based method, as it achieves the best performance in Table \ref{tabs:imgtranslation} apart from AsymFusion. The asymmetric method (AsymFusion) contains both shuffle and shift operations. Regions with sharp predication differences are indicated with red frames.}}
\label{pic:multimodal}
\end{figure*}

\textbf{Image Translation.} We benchmark on the dataset Taskonomy for performing image translation tasks, largely due to its data variety. In this part, we consider a wide range of modalities including depth, normal, shade, texture and edge, and aim to translate these data to RGB. Since there are no published methods of multimodal image translation that has such a data variety to date, we implement two commonly used fusion methods (concatenation and average) and two recently proposed fusion methods (attention and MMF). These four methods are treated as baselines for comparison. we adopt Fr$\acute{\text{e}}$chet Inception Distance (FID) score as the evaluation metric. FID compares the statistics of generated images against real images, by fitting a Gaussian distribution to the hidden activations of InceptionNet for compared images and compute Wasserstein-2 distance between these distributions. A lower FID score is better, indicating the generated images are more similar to real counterparts. In Table \ref{tabs:imgtranslation}, we provide results comparison for cases with two modalities and three modalities. For different combinations of modalities, our fusion models are  consistently better than other methods. Results also indicate that our method can be extended to more modalities and maintain competitive performance. In Figure \ref{pic:multimodal}, we provide illustrative examples for comparing different predicted results, where our proposed method outperforms the others.

\begin{table}[t]
\centering

\caption{\textmd{Semantic segmentation results comparison on NYUDv2 dataset. Baseline methods mostly adopt RefineNet as the unimodal network, except for SCN. For comparison convenience, we also apply our fusion methods on RefineNet, with two backbone architectures ResNet101 and ResNet152 respectively. We report pixel accuracy (\%), mean accuracy (\%) and mean IoU (\%).}}

\resizebox{83mm}{!}{
\begin{tabular}{ccc|ccc|c}
\toprule
Method & Data modality & Backbone &Pixel acc. & Mean acc.&  IoU & \#Params. \\
\midrule
\midrule
RefineNet \cite{lin2019refinenet}&RGB&ResNet101&73.8&58.8&46.4&118.10M\\
RefineNet \cite{lin2019refinenet}&RGB&ResNet152&74.4&59.6&47.6&133.74M\\
CFN \cite{DBLP:conf/iccv/LinCCHH17} &RGB-D& ResNet152&-&-&47.7&-\\
SCN \cite{DBLP:journals/tcyb/LinZJLH20} &RGB-D&ResNet152&-&-&49.6&-\\
RDFNet \cite{DBLP:conf/iccv/LeePH17} &RGB-D&ResNet101&75.6&62.2&49.1&366.71M\\
RDFNet \cite{DBLP:conf/iccv/LeePH17} &RGB-D&ResNet152&76.0&62.8&50.1&398.00M\\
\midrule
RefineNet $\dag$&RGB&ResNet101&73.8&59.0&46.5&118.10M\\
RefineNet $\dag$&Depth&ResNet101&64.0&45.6&34.3&118.10M\\
\textbf{AsymFusion} &RGB-D&ResNet101&{76.6}&{63.5}&{50.8}&\textbf{118.20M}\\
\textbf{AsymFusion} &RGB-D&ResNet152&{\textbf{77.0}}&{\textbf{64.0}}&{\textbf{51.2}}&\textbf{133.89M}\\
\bottomrule
\end{tabular}}
  \begin{tablenotes}
    \footnotesize
    \item[1] $\dag$ indicates our re-implemented results
     \end{tablenotes}
\vskip-0.5em
\label{tabs:nyudv2}

\end{table}

\begin{table}[t]
\centering

\caption{\textmd{Semantic segmentation results on Cityscapes dataset using 19 semantic labels. We apply our fusion methods to DeepLabv3+ (Xecption65). We include recently proposed top-performing methods for comparison. Extra data indicates whether use additional training dataset with coarse annotations for further improving performance. We report mean IoU (\%) as the evaluation metric.}}
\resizebox{83mm}{!}{
\begin{tabular}{cccc|p{1cm}<{\centering}|c}
\toprule
Method & Data modality & Extra data & Backbone &IoU & \#Params. \\
\midrule
\midrule
PSPNet \cite{DBLP:conf/cvpr/ZhaoSQWJ17}&RGB&$\times$&ResNet101&80.9&56.27M\\
DeepLabv3 \cite{DBLP:journals/corr/ChenPSA17} &RGB&$\times$&ResNet101&79.3&58.16M\\
Mapilary \cite{DBLP:conf/cvpr/BuloPK18}&RGB&$\times$& WideResNet38&78.3&135.86M\\
DeepLabv3+ \cite{DBLP:conf/eccv/ChenZPSA18}&RGB&$\times$&Xecption65&78.8&43.48M\\
DPC \cite{DBLP:conf/nips/ChenCZPZSAS18} &RGB&$\times$&Xecption65&80.9&41.82M\\
DRN \cite{DBLP:conf/icip/ZhuangYTMZLJX018}&RGB&$\times$&WideResNet38&79.7&129.16M\\
AdapNet++ \cite{DBLP:journals/ijcv/RussakovskyDSKS15}&RGB&$\surd$&ResNet50&81.2&30.20M\\
SSMA \cite{DBLP:journals/ijcv/RussakovskyDSKS15}&RGB-D&$\surd$&ResNet50&82.2&56.44M\\
\midrule
DeepLabv3+ $\dag$&RGB&$\times$&Xecption65&79.4&43.48M\\
DeepLabv3+ $\dag$&Depth&$\times$&Xecption65&62.3&43.48M\\
\textbf{AsymFusion} &RGB-D&$\times$&Xecption65&\textbf{82.1}&\textbf{43.52M}\\
\bottomrule

\end{tabular}}
  \begin{tablenotes}
    \footnotesize
    \item[1] $\dag$ indicates our re-implemented results
  \end{tablenotes}
  
\vskip-0.5em
\label{tabs:city}

\end{table}

\begin{table}[t]
\centering

\caption{\textmd{Image translation results comparison on Taskonomy dataset, under different combinations of modalities (two or three modalities) to verify the generalization of our models. All experiments adopt the same fusion layers as described in Section \ref{sec:imple}. FID score is used as the evaluation metric, the lower the better.} }
\vskip-0.3em
\resizebox{83mm}{!}{
\begin{tabular}{c|p{1.1cm}<{\centering}|p{1.1cm}<{\centering}|p{1.2cm}<{\centering}|p{1.1cm}<{\centering}|p{1.7cm}<{\centering}}
\toprule
Data modality & \makecell{Concat}& Average& Attention &MMF&\textbf{AsymFusion}\\
\midrule
\midrule
Shade,Depth&96.5&101.3&87.3&92.0&\textbf{82.5}\\
Normal,Texture&88.9&93.0&83.3&85.9&\textbf{77.8}\\
\midrule
Depth,Texture,Normal&86.4&90.2&81.5&82.1&\textbf{75.1}\\
Shade,Normal,Edge&92.8&94.4&85.6&88.6&\textbf{79.4}\\

\bottomrule
\end{tabular}}
\vskip-0.4em
\label{tabs:imgtranslation}

\end{table}

\subsection{Ablation Studies}
\label{discuss}
\textbf{Importance of Using Private BNs.} A quick question is that, what is the benefit of privatizing BNs? In Table \ref{tabs:sharing}, we compare three types of parameter sharing strategies in encoder. We adopt RefineNet (ResNet101), and conduct comparison experiments on NYUDv2 dataset. We obtain two conclusions from the results. Firstly, sharing BNs will lead to an obvious performance drop (4.5\% absolute drop for IoU). Secondly, compared with using individual networks, sharing convolutional parameters and leaving BNs privatized brings even slightly higher performance, yet largely reduces total parameters. Such finding would be instructive for multimodal learning.

\textbf{Components Analysis of AsymFusion.} We mainly design AsymFusion using two components, channel shuffle and pixel shift. Besides, we apply knowledge distillation for additional performance improvements, mentioned in Section \ref{sec:imple}. In this part, we verify the importance of these parts. Results shown in Table \ref{tabs:shuffleshift} indicate that shuffle and shift both play importance roles for the fusion performance, which together bring over 3\% IoU improvements. The knowledge distillation process further shows slight effects, with about 0.3\% IoU improvements. 

\begin{figure}[t]
\centering
\includegraphics[scale=0.34]{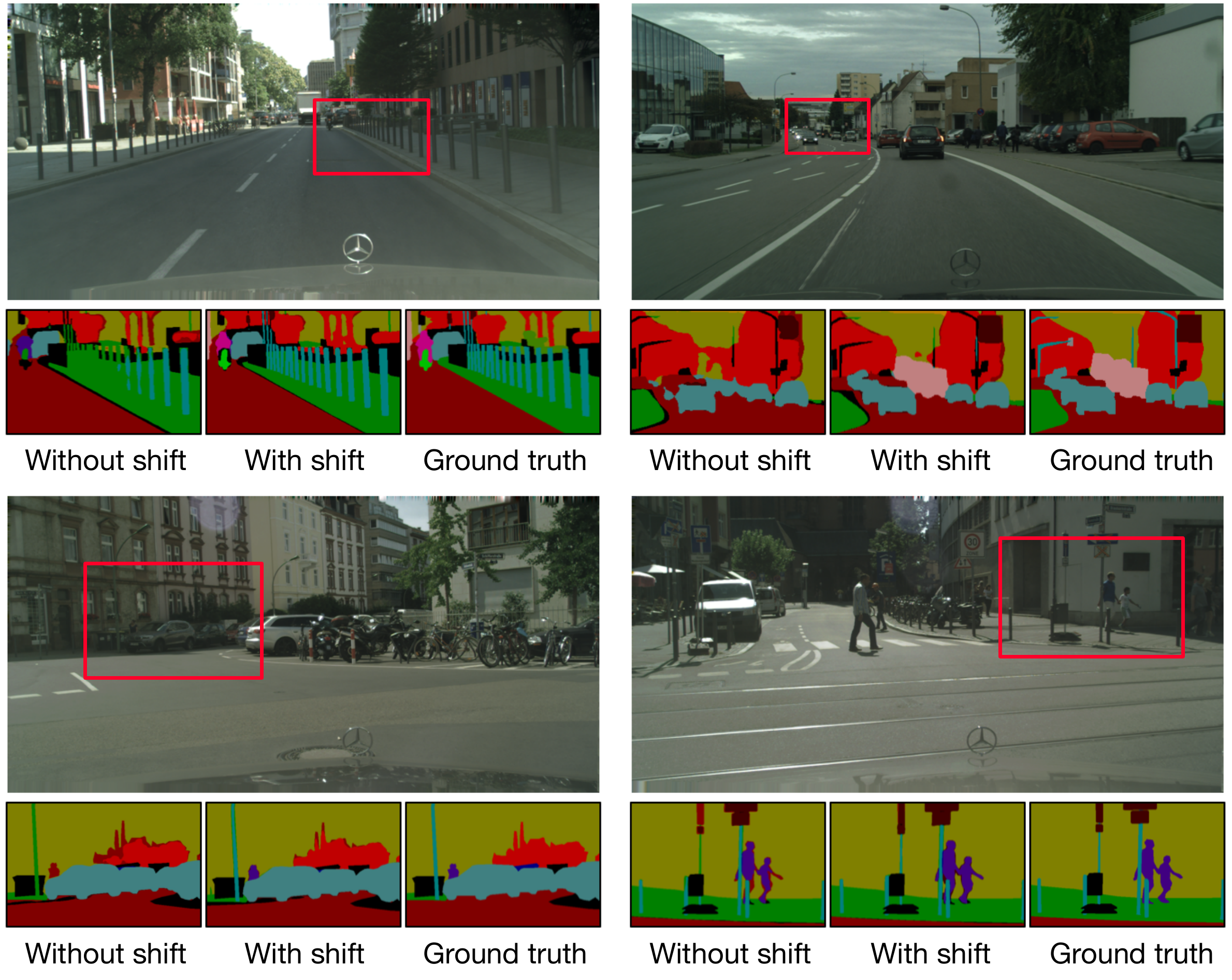}
\vskip-0.1em
\caption{\textmd{Illustrative semantic segmentation results on Cityscapes dataset, for comparing predictions without and with shift operations. Depth inputs are also used for prediction but are not shown. Red frames indicate regions for prediction, and prediction errors are highlighted in each prediction image. Results show that shift operations may benefit the prediction at small and distant objects.}}
\label{pic:detail}
\vskip-1.5em
\end{figure}

\textbf{Benefits of Using Shift Operations.} In Figure \ref{shuffle_shift}, we describe how to integrate pixel shift operations into residual blocks. When integrating shift operations, there are two additional skip connections across two branches. A question arises, is the improvement actually brought by these additional skip connections instead of shift operations? To figure it out, we keep these skip connections and only remove shift operations. The numerical results are 75.5\%/62.7\%/49.4\%, much lower than using shift operations (76.4\%/63.1\%/50.5\%, see Table \ref{tabs:shuffleshift}).
Illustrative results of predictions without and with shift operations are shown in Figure \ref{pic:detail}. By comparison, predictions with shift operations tend to be better at capturing details, including thin and small objects, e.g., poles, distant persons and vehicles.

\vskip0.2em 
\textbf{Comparison of Fusion Directions.} In Table \ref{tabs:direction}, we verify that for multi-layer fusion, the direction of fusion has impact on the fusion performance. For all reported fusion methods, we observe that fusion from depth branch to RGB branch shows better results than fusion with the opposite direction. Applying symmetric fusion methods to the bidirectional scheme shows minor drops than unidirectional counterparts sometimes. We conjecture that symmetric fusion methods tend to have similar feature representation and thus are not very compatible with bidirectional fusion, as described in Section \ref{sec:shuffleshift}. Our proposed asymmetric fusion method presents large performance gains when using bidirectional fusion.

\begin{table}[t]
\centering

\caption{\textmd{Exploration of different parameter sharing strategies for the encoder, based on NYUDv2 and Cityscapes datasets. We compare three strategies including using individual networks (widely adopted by existing multimodal works), sharing convolutional parameters and BNs, and sharing only convolutional parameters with individual BNs. We report pixel accuracy (\%) and mean IoU (\%).}}
\resizebox{83mm}{!}{
\begin{tabular}{ll|p{1.3cm}<{\centering}p{1cm}<{\centering}|c}
\toprule
Dataset&$\;\;\;\;\;\;$Parameter sharing strategy & Pixel acc. &IoU & \#Params. \\
\midrule
\midrule
\multirow{3}*{NYUDv2}&Individual Convs + Individual BNs&76.1&50.5&236.20M\\
&Shared Convs $\;\;\;\,$ + Shared BNs&72.2&46.3&118.10M\\
&Shared Convs $\;\;\;\,$ + Individual BNs&\textbf{76.6}&\textbf{50.8}&118.20M\\
\midrule
\multirow{3}*{Cityscapes}&Individual Convs + Individual BNs&96.8&81.9&87.04M\\
&Shared Convs $\;\;\;\,$ + Shared BNs&95.3&78.7&43.48M\\
&Shared Convs $\;\;\;\,$ + Individual BNs&\textbf{97.0}&\textbf{82.1}&43.52M\\
\bottomrule
\end{tabular}}
\label{tabs:sharing}

\end{table}

\begin{table}[t]
\centering

\caption{\textmd{Comparison different components of our proposed fusion methods, containing channel shuffle, pixel shift, and the knowledge distillation applied at the end of the network. Experiments are conducted with RefineNet (ResNet101) on NYUDv2 dataset. We report pixel accuracy (\%), mean accuracy (\%) and mean IoU (\%).}}
\resizebox{83mm}{!}{
\begin{tabular}{p{0.9cm}<{\centering}p{0.9cm}<{\centering}p{1.4cm}<{\centering}|p{1.4cm}<{\centering}p{1.4cm}<{\centering}p{1.2cm}<{\centering}|c}
\toprule
Shuffle & Shift & Distillation & Pixel acc. & Mean acc.&  IoU & \#Params. \\
\midrule
\midrule
$\times$&$\times$&$\times$&74.0&58.9&47.3&118.20M\\
$\times$&$\times$&$\surd$&74.3&59.3&47.6&118.20M\\
\midrule
$\surd$&$\times$&$\times$&75.2&62.5&49.3&118.20M\\
$\times$&$\surd$&$\times$&74.8&61.7&48.7&118.20M\\
$\surd$&$\surd$&$\times$&{76.4}&{63.1}&{50.5}&118.20M\\
\midrule
$\surd$&$\surd$&$\surd$&{76.6}&{63.5}&{50.8}&118.20M\\
\bottomrule
\end{tabular}}
\label{tabs:shuffleshift}

\end{table}

\begin{table}[t]
\centering

\caption{\textmd{Results comparison of symmetric fusion methods and the proposed asymmetric fusion method. We compare two settings, i.e., using individual / shared convolutional parameters. Note that individual BNs are adopted for all experiments in this table. Symmetric fusion methods include concatenation, average and attention. The fusion operation addition is omitted as it performs likely to the average operation. We report pixel accuracy (\%) / mean IoU (\%).}}
\resizebox{83mm}{!}{
\begin{tabular}{p{1.3cm}<{\centering}|c|p{1.3cm}<{\centering}|p{1.3cm}<{\centering}|p{1.3cm}<{\centering}|p{1.75cm}<{\centering}}
\toprule
Params. &Fusion direction & \makecell{Concat}& Average& Attention &\textbf{AsymFusion}\\
\midrule
\midrule

\multirow{3.4}*{Individual}
&Depth$\to$RGB&74.9 / 48.6&74.7 / 48.2&75.3 / 49.1&75.2 / 49.3\\
&RGB$\to$Depth&74.0 / 47.1&73.6 / 46.6&74.6 / 48.5&74.9 / 48.7\\
\cmidrule[0.4pt](r){2-6}
&Bidirectional&74.7 / 48.0&74.2 / 47.7&74.9 / 48.7&\textbf{76.2 / 50.5}\\

\midrule
\multirow{3.4}*{Shared}
&Depth$\to$RGB&75.3 / 48.9&74.9 / 48.5&75.5 / 49.4&75.4 / 49.4\\
&RGB$\to$Depth&74.2 / 47.3&73.9 / 46.8&74.8 / 48.6&75.2 / 49.1\\
\cmidrule[0.4pt](r){2-6}
&Bidirectional&75.0 / 48.4&74.6 / 48.2&75.1 / 48.9&\textbf{76.6 / 50.8}\\

\bottomrule
\end{tabular}}
\label{tabs:direction}

\end{table}

\section{Conclusion}
\vskip0.3em
We present a compact and effective multimodal fusion framework. To start, we propose that multimodal training can be realized in one single network with modality-specific BNs, enabling implicit fusion via joint feature representation training. Regarding fusion methods, to make advantage of the bidirectional fusion scheme, we propose channel shuffle and pixel shift operations that are asymmetric with respect to fusion directions. These operations strengthen the interaction of multimodal information flow, and tend to improve feature representation ability to discriminate rich edge-aware information from context. Experimental results on several tasks indicate that our fusion scheme outperforms state-of-the-art counterparts, with only about 0.1\% additional parameters than a given unimodal network.

\begin{spacing}{0.55}
\end{spacing}

\section*{Acknowledgement}
\vskip0.3em
This work is  jointly supported by the National Science Foundation of China (NSFC) and the German Research Foundation (DFG) in project Cross Modal Learning, NSFC 61621136008/DFG TRR-169.

\newpage
\bibliographystyle{splncs04}
\balance
\bibliography{egbib}

\end{document}